\documentclass[11pt]{article}

\usepackage[final]{acl}

\usepackage{times}
\usepackage{latexsym}

\usepackage[T1]{fontenc}

\usepackage[utf8]{inputenc}

\usepackage{microtype}

\usepackage{inconsolata}

\usepackage{graphicx}

\usepackage{soul}
\usepackage[normalem]{ulem}
        
\title{Evaluating the Role of Verifiers in Test-Time Scaling for Legal Reasoning Tasks}

\author{Davide Romano, Jonathan Richard Schwarz, Daniele Giofrè \\
        Thomson Reuters \\ \texttt{\{Davide.Romano2, Jonathan.Schwarz, Daniele.Giofre\}@thomsonreuters.com}}
        
\begin{document}
\maketitle

\begin{abstract}
Test-time scaling (TTS) techniques can improve the performance of large language models (LLMs) at the expense of additional computation and latency. While TTS has proven effective in formal domains such as mathematics and programming \citep{snell2024scaling, chen2024more}, its value in argumentative domains such as law remains underexplored. We present an empirical study of verifier-based TTS methods for legal multiple-choice QA (MCQA) across five benchmarks. Using a family of 7 reward models, we evaluate both outcome-level (Best-of-$N$) and process-level (tree search) verification under realistic low-$N$ budgets. Our analysis systematically investigates how verifier utility is affected by key properties such as domain specialization, model size, and supervision type (process-supervised PRMs vs. outcome-only ORMs), even when applied across different roles.

\end{abstract}

\section{Introduction}
Test-Time Scaling (TTS) methods aim to enhance Large Language Model (LLM) performance by trading additional compute for improved accuracy at inference time. The broad spectrum of these techniques range from single-path approaches like generating longer Chains-of-Thought (CoT) \citep{wei2022chain, guo2025deepseek, jaech2024openai} to more complex parallel and verifier-guided methods such as Best-of-N (BoN) selection and tree search. 
\\
Systematic investigations of these verifier-guided methods in formal domains like math and programming have demonstrated substantial accuracy improvements on multiple choice QA (MCQA) tasks \citep{brown2024large, wu2024inference, snell2024scaling}. However, the legal domain presents distinct challenges; its reasoning is often defeasible and accommodates multiple valid analytical paths. While prior work has explored single-path inference for legal reasoning \citep{yu2025evaluating}, investigations into parallel and verifier-based TTS are notably absent from the literature. 
This gap is critical because the verifiers underpinning these methods are often trained on general-purpose or formal-domain data.
It remains an open question whether such models can reliably evaluate legal reasoning, or if domain-specific verifiers are required to achieve meaningful gains.
These verifiers are broadly categorized into two types: Outcome Reward Models (ORMs), which assign a single score to a complete output, and Process Reward Models (PRMs), which provide fine-grained, step-by-step feedback \citep{uesato2022solving, lightman2023let}. This paper addresses the aforementioned gap by empirically investigating whether the performance enhancements observed in formal domains translate to legal MCQA. Through an extensive comparison of reward models, we analyze how to optimize these verification strategies by evaluating the impact of verifier domain specialization, model size, and supervision type (PRM vs. ORM).

\begin{figure}[t]
  \includegraphics[width=\columnwidth]{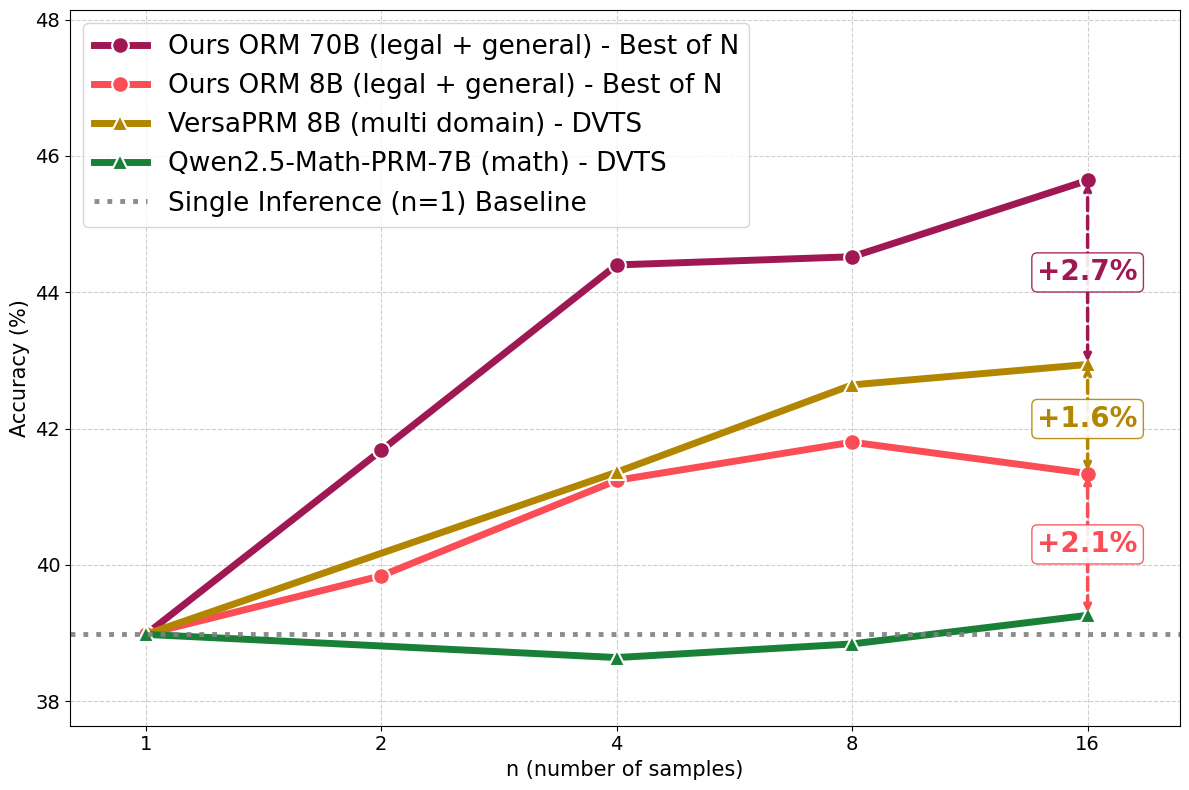}
  \caption{TTS with Llama-3.1-8B-Instruct with four different verifiers from N=4 to N=16, average over 5 legal MCQA benchmarks}
  \label{fig:experiments}
\end{figure}

\noindent We examine verifier-based TTS for legal reasoning to answer the following research questions:
\begin{itemize}
    \item \textbf{RQ1 (Value of verification under matched compute).} Under equal compute budgets, do outcome-verified Best-of-$N$ (BoN) and Diverse Verifier Tree Search (DVTS) outperform simple Majority Vote (MV) on legal MCQA benchmarks?
    \item \textbf{RQ2 (Importance of domain specialization and verifier size).} With method and compute held constant, does a legal-specialized verifier outperform a general-domain verifier, and how large is the additional effect of scaling the verifier size?
    \item \textbf{RQ3 (Role transfer between verifiers)}: Under matched size and compute, could these reward models be used also out-of-role (for PRMs in outcome verification and for ORMs in process-verification) in legal MCQA tasks?
\end{itemize}

\paragraph{Contributions} This paper makes three key contributions to the understanding of verifier-based test-time scaling for legal reasoning. First, we conduct a comprehensive comparison between MV, BoN, and process-verified DVTS using open-source models, revealing that verifier-based methods rarely outperform simple voting baselines by significant margins in legal reasoning. Second, through systematic ablation studies, we show that both verifier model size and domain specialization are crucial for improving performance, with legal-domain training providing a distinct advantage that becomes most apparent at larger scales. Notably, we find that the utility of all methods diminishes as generator model capability increases, with even sophisticated verification providing minimal gains. Finally, our analysis of supervision type shows that PRMs deliver superior performance compared to ORMs of similar size, even when PRMs are deployed outside their intended role for outcome verification tasks. These findings provide valuable guidance for practitioners seeking to optimize computational resources in legal NLP applications.

\begin{figure*}[h]
  \includegraphics[width=0.94\linewidth]{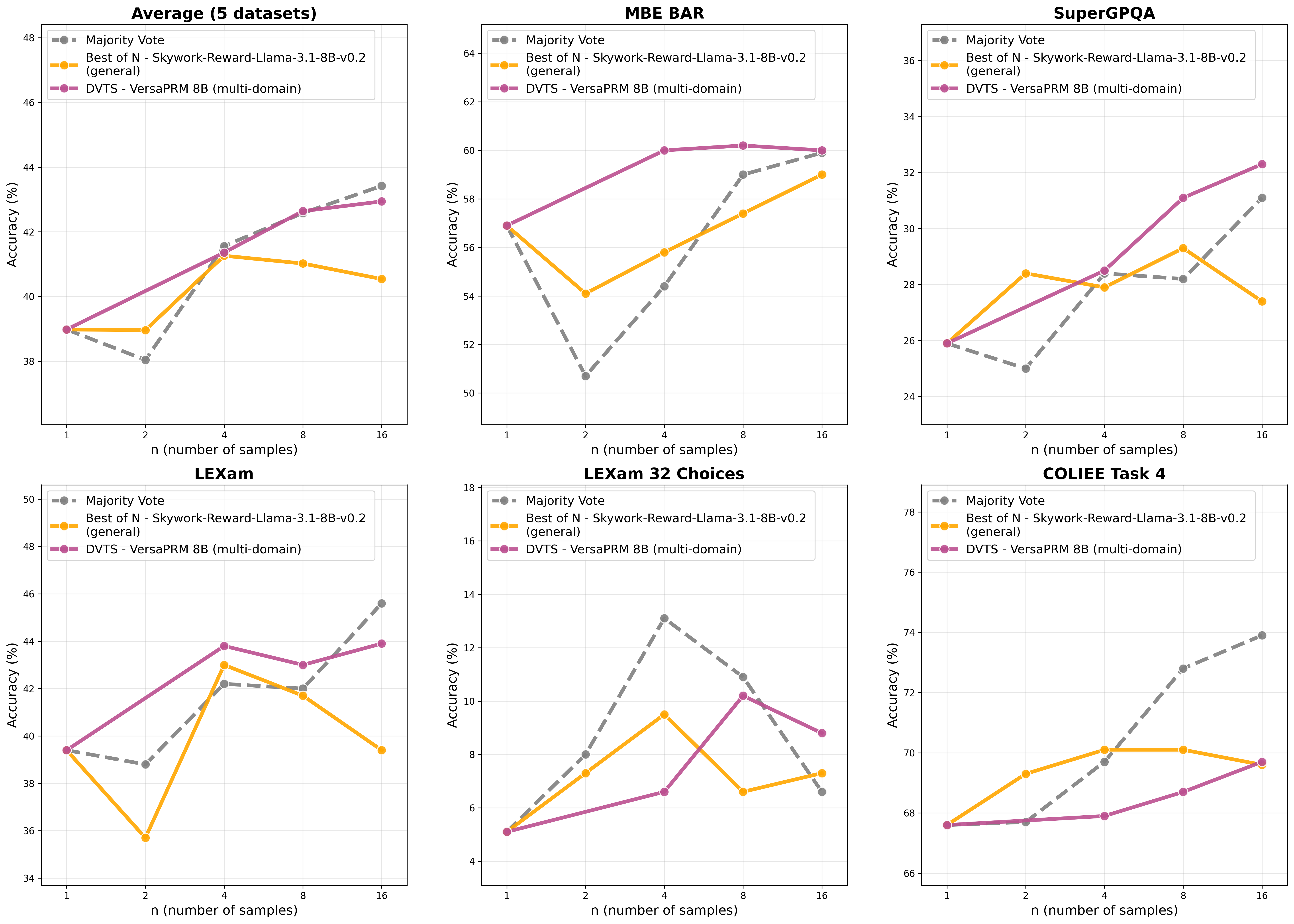}
  \caption {RQ1 results across all benchmarks with \texttt{Llama 8B} as the generator}
  \label{fig:rq1}
\end{figure*}

\section{Experimental Setup}

We test three generators: \textbf{Llama-3.2-3B-Instruct}, \textbf{Llama-3.1-8B-Instruct}, and \textbf{Llama-3.1-70B-Instruct} \citep{dubey2024llama}. We ran our tests with CoT prompting \cite{wei2022chain} and temperature $T=0.8$ and the system prompts in appendix \ref{sec:appendix-prompt}. Our evaluation compares three methods:

\noindent\textbf{Majority Vote (MV):} Sample $k$ CoT responses and select the most frequent answer.

\noindent \textbf{Best-of-$N$ (BoN):} Sample $N$ CoT responses, score each with an Outcome Reward Model (ORM), and select the one with highest reward.

\noindent \textbf{DVTS \citep{beeching2024scaling}:} A tree search guided by a Process Reward Model (PRM) scoring partial steps. For optimal hyperparameter choice, such as expansion width or aggregation strategy, we ran an ablation study on MBE exam that can be found in appendix \ref{sec:appendix-tuning}.


While generating $N$ trajectories of length $T$ dominates computational cost at $\mathcal{O}(T^2)$, the reward-model scoring methods (Best-of-N and DVTS) add only modest linear $\mathcal{O}(T)$ overhead, making all three approaches comparable in runtime (see more details in Appendix \ref{sec:appendix-compute}) .

\noindent We use the verifiers detailed in Table~\ref{tab:verifiers} and evaluate on five legal benchmarks: binary \textbf{COLIEE Task 4} \citep{coliee2025benchmark}; four-option \textbf{MBE} BAR Exam and \textbf{LEXam} \citep{fan2025lexam}; eight-option \textbf{SuperGPQA (Law subset)} \citep{du2025supergpqa}; and thirty-two-option \textbf{LEXam-32}\citep{fan2025lexam}.
\textbf{MBE} BAR Exam is the only restricted-access benchmark.

\begin{table}[h]
\centering
\small
\begin{tabular}{l p{2.4cm} l l}
\hline
\textbf{Type} & \textbf{Verifier Model} & \textbf{Size} & \textbf{Training data} \\
\hline
ORM & \texttt{Our RMs} & 8B, 70B & General + Legal$^\dagger$ \\
 & \texttt{Skywork-Reward \citep{liu2024skywork}} & 8B, 27B & General \\
\hline
PRM & \texttt{VersaPRM} \citep{zeng2025versaprm} & 8B & Multi-domain$^*$ \\
 & \texttt{Qwen2.5-Math-PRM \citep{prmlessons}} & 7B, 72B & Math \\
\hline
\end{tabular}
\caption{Verifiers used in our study, grouped by supervision type (ORM/PRM). 
$^*$Multi-domain includes Law, Philosophy,
Biology and others. 
$^\dagger$Both our models were trained on identical datasets, comprising general knowledge sources such as UltraFeedback \cite{cui2023ultrafeedback} and restricted-access legal data from US and UK jurisdictions. This training corpus encompasses various task types: legal reasoning, legal information retrieval, legal summarization, and basic instruction following.}
\label{tab:verifiers}
\end{table}
\subsection{Study Design}

\paragraph{RQ1 (Value of Verification):} We compare MV against both BoN with \texttt{Skywork-RM-27B}, and DVTS (with \texttt{VersaPRM-8B}). For BoN, we use \texttt{Skywork-RM-27B} due to its performance on the RewardBench \cite{lambert2024rewardbench} benchmark. For DVTS, we selected VersaPRM-8B as it is the first open-source, multi-domain PRM available for research.
\textbf{RQ2 (Impact of Domain \& Size):} We compare legal (\texttt{Ours RMs}) vs. general (\texttt{Skyworks}) ORMs using BoN, and the multi-domain (\texttt{VersaPRM}) vs. out-of-domain (\texttt{Qwen-Math-PRM}) PRMs using DVTS.
\textbf{RQ3 (Supervision Type):} We test our legal ORMs (\texttt{Ours}) against the legal PRM (\texttt{VersaPRM}) on both BoN and DVTS.

\begin{figure*}[h]
  \includegraphics[width=1\linewidth]{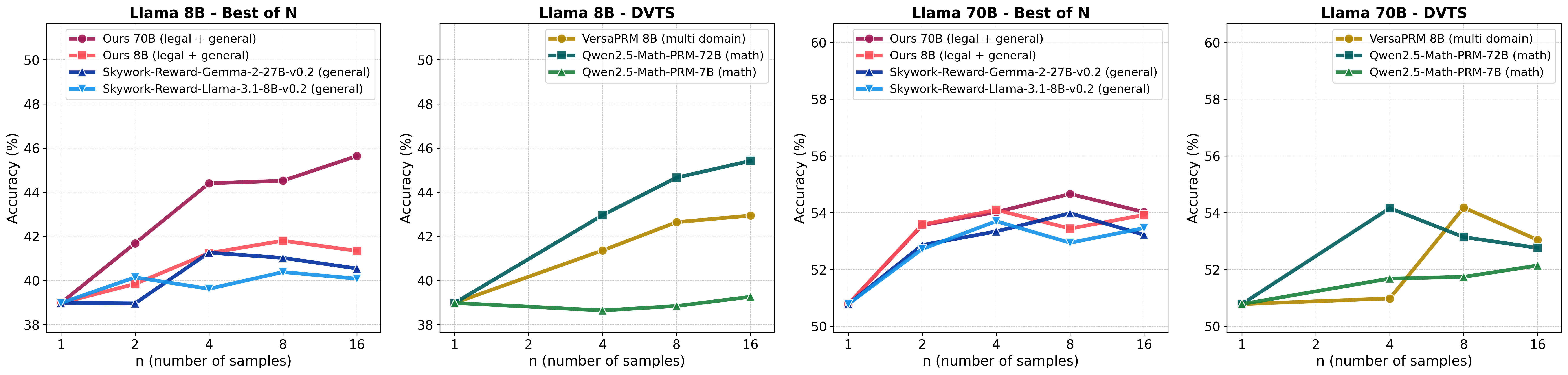}
  \caption {RQ2 average results with both \texttt{Llama 8B} and \texttt{Llama 70B}}
  \label{fig:rq2}
\end{figure*}

\begin{figure*}[h]
  \includegraphics[width=1\linewidth]{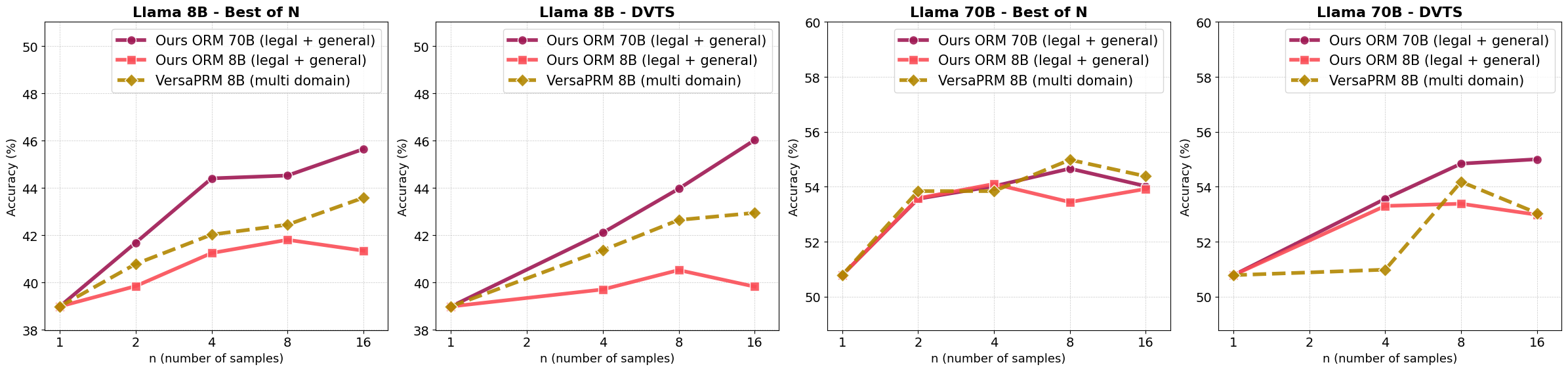}
  \caption {RQ3 average results with both \texttt{Llama 8B} and \texttt{Llama 70B}}
  \label{fig:rq3}
\end{figure*}

\section{Results}

\label{sec:results}

\subsection{RQ1: Value of verification under matched compute}
\label{sec:rq1}
In Figure~\ref{fig:rq1} and figures \ref{fig:rq1-llama3}-\ref{fig:rq1-llama70b} in appendix \ref{appendix:full-results} we can see that MV remains a strong baseline across benchmarks and generator scales. The only model where BoN and DVTS surpass on average across the 5 benchmarks the MV baseline is \texttt{Llama-3.2-3B-Instruct}, where they achieve around 1.4\% average improvement. For larger models, however, verifications provides no performance benefit or even show a decrease compared to MV. These limited benefits of verification prompted us to further explore the potential of domain specialization and size as two variables to obtain better verification in \ref{sec:rq2}.

\begin{table}[h]
    \centering
    \caption{Relative performance Gains on Llama 70B Generator at N=16 against Majority Vote baseline. Best-of-N uses Legal ORM 70B and DVTS uses QwenPRM 72B.}
    \label{tab:gains_llama70b}
    \resizebox{0.8\columnwidth}{!}{%
    \begin{tabular}{@{}l|l|c@{}}
        \toprule
        \textbf{Benchmark} & \textbf{Method} & \textbf{Rel. Gain} \\
        \midrule
        \multirow{2}{*}{MBE BAR} & Best-of-N & +0.6 \\
         & DVTS & \textcolor{red}{-4.8} \\
        \midrule
        \multirow{2}{*}{SuperGPQA} & Best-of-N & \textcolor{red}{-2.1} \\
         & DVTS & \textcolor{red}{-4.7} \\
        \midrule
        \multirow{2}{*}{LEXam} & Best-of-N & \textcolor{red}{-1.3} \\
         & DVTS & \textcolor{red}{-0.5} \\
        \midrule
        \multirow{2}{*}{LEXam 32} & Best-of-N & +10.2 \\
         & DVTS & \textbf{+12.4} \\
        \midrule
        \multirow{2}{*}{COLIEE Task 4} & Best-of-N & \textcolor{red}{-1.4} \\
         & DVTS & \textcolor{red}{-2.7} \\
        \bottomrule
    \end{tabular}%
    }
\end{table}

\paragraph{Performance Variation Across Benchmarks} A closer analysis of the per-benchmark performance reveals that the utility of verification is heavily influenced by the task's complexity, specifically the cardinality of the answer space. While this is more limited with smaller generators, with 70B generator as shown in Table \ref{tab:gains_llama70b} the difference is considerable. On benchmarks with a small number of answer choices, such as COLIEE Task 4, MBE, and LEXam, verifier-based methods offer marginal or even negative gains over the highly effective MV baseline. However, this trend is starkly reversed in the high-complexity \textbf{LEXam-32} task. With 32 possible answers, the output space becomes significantly noisier, causing the MV baseline to struggle. In this scenario, DVTS achieves a substantial relative gain of \textbf{+12.4\%}.

\subsection{RQ2: Domain specialization vs.\ verifier size}
\label{sec:rq2}

Figure~\ref{fig:rq2} and figures \ref{fig:rq2_bar}-\ref{fig:rq2_superg} in appendix \ref{appendix:full-results} demonstrate that BoN with \emph{our} reward models (both 8B and 70B variants) match or outperform general-domain verifiers across evaluations.
While the 8B model shows minimal performance advantages over general verifiers, the 70B model consistently delivers superior results across numerous benchmarks. 
Regarding the performance of PRMs with DVTS, \texttt{QwenPRM 72B} produces the most significant enhancement when coupled with smaller generator models. Direct comparison between similarly sized \texttt{VersaPRM 8B} and \texttt{QwenPRM 7B} reveals that \texttt{VersaPRM} consistently delivers superior performance. 



\subsection{RQ3: Supervision type and transfer (PRM vs.\ ORM)}
\label{sec:rq3}
Figure~\ref{fig:rq3} and figures \ref{fig:rq3_bar}-\ref{fig:rq3_superg} in Appendix \ref{appendix:full-results} show that PRMs provide consistent benefits: as BoN scorers they yield stronger reranking than size-matched ORMs, and within DVTS they offer more effective guidance. Improvements are concentrated on the smaller generators, similar to the other results. \texttt{Ours 70B} can still perform better than \texttt{VersaPRM} also in process supervision even though it has received no process training.

\subsection{Discussion}

\begin{table}[h]
    \centering
    \caption{Relative improvement values across Llama models at N=16 over the Majority Vote baseline at N=16.}
    \label{tab:average_gains_comparison}
    \resizebox{\columnwidth}{!}{%
    \begin{tabular}{@{}l|c|c|c@{}}
        \toprule
        {\textbf{Method + Reward Model}} & \textbf{Llama 3B} & \textbf{Llama 8B} & \textbf{Llama 70B} \\
        \midrule
        BoN + VersaPRM 8B & +2.94 & +0.16 & \textbf{+1.56} \\
        BoN + Legal ORM 70B & \textbf{+4.46} & \textbf{+2.22} & +1.20 \\
        DVTS + QwenPRM 72B & +4.00 & +2.00 & \textcolor{red}{-0.06} \\
        \bottomrule
    \end{tabular}%
    }
\end{table}

\paragraph{Diminishing Returns of Verification} The performance gains from verifier-based TTS decrease as the capability of the generator model improves. At same value of $N$ when using the 70B generator, even well-configured verifiers provide only small improvements over the MV baseline (Table \ref{tab:average_gains_comparison}), which proves to be a very competitive method.

\paragraph{Task Complexity as a Key Differentiator} The strong performance on the LEXam-32 benchmark provides a crucial insight into the practical limits of MV method. While MV is a robust baseline for tasks with a small set of discrete answers, its utility appears to degrade significantly as the solution space expands. It is in this high-cardinality environment that verifier-guided methods demonstrate their value. This suggests that for problems with a constrained output space, the simplicity of MV may be sufficient, but as task complexity and the number of possible outcomes grow, the computational overhead of verification becomes a more justifiable investment. This relationship warrants further investigation, including in Open QA settings.



\paragraph{The Dual Impact of Scale and Specialization}  Our findings highlight two key drivers of verifier performance: model scale and domain specialization. Scaling a verifier from an 8B to a 70B model consistently yields substantial performance gains. Similarly, models trained on specialized legal data regularly outperform their general-domain counterparts. However, these two factors are linked. The advantage from specialization is most pronounced at the 70B scale, while at the 8B scale, specialized models like VersaPRM offer modest, yet clear, improvements. This indicates that while larger models are inherently more capable, targeted training provides a distinct, scale-dependent advantage.

\paragraph{The Generalization of Process Supervision} Finally, we find that PRMs seem to work well also in both outcome and process verification for legal tasks. This may indicate that the step-by-step feedback used to train PRMs helps them develop a more robust measure of reasoning quality.

\section{Conclusions}
In this work, we presented a systematic evaluation of verifier-based TTS for legal multiple-choice question answering. 


A consistent observation is the diminishing return of verification as the generator model's power increases; while gains are evident for smaller generators, they shrink significantly for more powerful ones, where a simple Majority Vote often remains competitive. However, the utility of Majority Vote is challenged by task complexity, and we find that verifier-based methods provide substantial gains in high-cardinality benchmarks where the answer space is large.


Crucially, we find that effective verification relies on the dual impact of model scale and domain specialization. The advantage of in-domain training is most pronounced at larger verifier scales. This is complemented by the notable generalization of process supervision: \texttt{VersaPRM} proved highly versatile, outperforming size-matched Outcome Reward Models even when used out-of-role for outcome reranking.


For practitioners, these findings suggest that investing in high-quality, in-domain reward models is a promising direction for improving inference-time legal reasoning.

\section{Limitations and Scope}
The scope of this study is limited to legal reasoning MCQA, and our findings may not generalize to other legal tasks such as summarization or open-ended QA where verification is arguably more complex. Additionally, our experiments primarily focused on a single model family (i.e. \texttt{Llama 3.1} and \texttt{Llama 3.2}), and other model architectures might exhibit different improvements from verification. Future work should explore additional legal domains and open QA, expand the verifier pool with more recent reward models such as \texttt{Skywork-v2} Reward Models, and evaluate newer generators like \texttt{Qwen3} models. It should be noted that some verifiers used in this study are restricted-access data (\emph{ours} RMs), which limits the full reproducibility of certain results.

\bibliography{custom}

\appendix
\label{sec:appendix}

\section{TTS methods compute cost}
\label{sec:appendix-compute}

For $N$ trajectories with average length $T$, generation dominates total cost. Without KV caching, the generator cost scales as $\Theta(P_M N T^2)$. In contrast, reward-model scoring is linear in ($T$): Best-of-N adds one verifier forward per trajectory ($\Theta(P_R N T)$), and DVTS scores multiple partial paths across $s$ reasoning steps, about $\tfrac{s+1}{2}$ times the BoN cost, $\Theta(P_R N T\tfrac{s+1}{2})$. With typical CoT lengths of hundreds of tokens (average is 1000 for our CoTs) and $s \approx10$, these $\mathcal{O}(T)$ verifier terms remain much smaller than the $\mathcal{O}(T^2)$ generation term, even when the verifier is larger than the generator, so for a fixed sample count $N$, MV, BoN, and DVTS have comparable runtime, with only modest linear overheads for BoN and DVTS.

\section{Generator prompt templates}
\label{sec:appendix-prompt}

This section details the prompt templates used for the generator models.

\begin{table*}[h]
\centering
\begin{tabular}{p{2.2cm}p{2.5cm}l l p{2cm} p{2cm} l}
\hline
\textbf{Generator}     & \textbf{Reward Model}         & \textbf{Dataset} & \textbf{N} & \textbf{Expansion Width} & \textbf{Independent Subtrees} & \textbf{Accuracy} \\ \hline
Llama-3.1-8B-Instruct  & VersaPRM             & BAR              & 16         & 8                        & 2                             & 55.6\%            \\
                       &                      & BAR              & 16         & 4                        & 4                             & 59.5\%            \\
                       &                      & BAR              & 16         & \textbf{2}               & \textbf{8}                    & \textbf{61.8\%}   \\
                       &                      &                  &            &                          &                               &                   \\ \hline
Llama-3.1-70B-Instruct & Ours 70B             & LEXam 32         & 16         & 4                        & 4                             & 19.7\%            \\
                       &                      & LEXam 32         & 16         & \textbf{2}               & \textbf{8}                    & \textbf{22.6\%}   \\
\textbf{}              &                      &                  & \textbf{}  & \textbf{}                &                               &                   \\
                       & VersaPRM             & LEXam 32         & 16         & 4                        & 4                             & \textbf{19.0\%}   \\
                       &                      &                  &            & \textbf{2}               & \textbf{8}                    & 21.2\%            \\
                       &                      &                  &            &                          &                               &                   \\
                       & Qwen2.5-Math-PRM-72B & LEXam 32         & 16         & \textbf{4}               & \textbf{4}                    & \textbf{20.4\%}   \\
                       &                      &                  &            & 2                        & 8                             & 19.0\%            \\
                       &                      &                  &            &                          &                               &                  
\end{tabular}
\caption{\label{table-expansion}
    Expansion width tuning tests
  }
\end{table*}

\subsection{Majority Vote \& Best-of-N system prompt}
\label{sec:appendix-bon-prompt}

For the Best-of-N (BoN) generation method, we use a custom system template for MCQA and the classic CoT preprompt.

\begin{tcolorbox}[colback=gray!10!white, colframe=black, boxrule=0.5pt, arc=2mm]
Please complete the following user request.
\\~\\
When answering questions, first reflect on the problem step by step. At the end ALWAYS conclude with this phrase:
\\~\\
Therefore, the final answer is: \verb|\boxed{answer}|. I hope it is correct.
\\~\\
Where answer CAN BE ONLY \verb|[answer_options]|.
\end{tcolorbox}

The \{answer\_options\} are specific for each dataset, and it represents the list of accepted final answers. For example for BAR:
\begin{tcolorbox}[colback=gray!10!white, colframe=black, boxrule=0.5pt, arc=2mm]
Where answer CAN BE ONLY ONE OF THE FOLLOWING: "A", "B", "C", "D"'
\end{tcolorbox}
\noindent For parsing we accepted both formats "\verb|\boxed{answer}|" and "Therefore, the final answer is: \{answer\}" as Llama3.1 family didn't output the \verb|\boxed{}| very often. When we do the selection of the final answer we filter for the ones that passed successfully the parsing.

\subsection{DVTS system prompt}
\label{sec:appendix-dvts-prompt}

\begin{tcolorbox}[colback=gray!10!white, colframe=black, boxrule=0.5pt, arc=2mm]
Please complete the following user request.
\\~\\
Use this step-by-step format:
\\~\\
\#\# Step 1\\
\verb|[Reasoning step description]|
\\~\\
\#\# Step 2\\
\verb|[Reasoning step description]|
\\~\\
...
\\~\\
Regardless of the approach, ALWAYS conclude with this phrase:
\\~\\
Therefore, the final answer is: \verb|\boxed{answer}|. I hope it is correct.
\\~\\
Where answer CAN BE ONLY \verb|[answer_options]|.
\end{tcolorbox}

\section{RMs hyperparameter tuning}
\label{sec:appendix-tuning}

\subsection{Expansion width tuning}

Diverse Verifier Tree Search (DVTS) \citep{beeching2024scaling} requires to set a hyperparameter called "expansion width" $W$ which corresponds to the number of next steps expansions for each tree. Together with it, we have the number of initial subtrees $T$ (or also called "beams") at the start of the algorithm.
The number $N$ used in our paper is the corresponding of $W\cdot T$. To study the best parameter for $W$ given a fixed budget $N$ we performed the experiments in Table \ref{table-expansion}.

\noindent The results indicate that with smaller models such as \texttt{Llama-3.1-8B-Instruct} having a smaller expansion width (and therefore higher number of subtrees $T$) leads to better results. This is caused by the fact that diversifying more the generations at the start will lead to less formatting errors to parse the final answer. Therefore, in all our experiments we used $W=2$ and the number $T$ is $N/W$.

\subsection{Score aggregation method tuning}

From \cite{beeching2024scaling, zeng2025versaprm} there are four common options to the aggregation strategy choice for the PRM scores:
\\
\textbf{Min-Aggregation}
\[
\operatorname{Aggr}_{\min}(S) = \min_{i \in [k]} \operatorname{PRM}(S)_i .
\]
\\
\textbf{Last-Aggregation}
\[
\operatorname{Aggr}_{\text{last}}(S) = \operatorname{PRM}(S)_k .
\]
\\
\textbf{Average-Aggregation} 
\[
\operatorname{Aggr}_{\text{avg}}(S) = \frac{1}{k} \sum_{i \in [k]} \operatorname{PRM}(S)_i .
\]
\textbf{Prod-Aggregation} 
\[
\operatorname{Aggr}_{\text{prod}}(S) = \prod_{i \in [k]} \operatorname{PRM}(S)_i .
\]

To select our selection strategy we ran tests for VersaPRM and \texttt{Qwen2.5-Math-PRM-72B} on the BAR exam with \texttt{Llama-3.1-8B-Instruct} as generator with $N=16$. The results are in Table \ref{table-aggregation}

\noindent These results brought us to use for \texttt{VersaPRM} the \textbf{Mean} aggregation strategy. While for the QwenPRMs (both 72B and 7B) to use \textbf{Last}.

\begin{table}[h]
\resizebox{\columnwidth}{!}{%
\begin{tabular}{p{2cm}p{2cm}l}
\hline
\textbf{PRM}         & \textbf{Aggregation Strategy} & \textbf{BAR accuracy} \\ \hline
VersaPRM             & \textbf{Mean}                 & \textbf{62.7\%}       \\
                     & Min                           & 62.2\%                \\
                     & Last                          & 59.5\%                \\
                     &                               &                       \\ \hline
Qwen2.5-Math-PRM-72B & Mean                          & 57.7\%                \\
                     & Prod                          & 59.2\%                \\
                     & Min                           & 57.6\%                \\
                     & \textbf{Last}                 & \textbf{60.5\%}      
\end{tabular}%
}
\caption{\label{table-aggregation}
    Aggregation strategy ablation tests
  }
\end{table}


\section{Full results}
\label{appendix:full-results}
Full results are added, I just commented them for faster compilation

\begin{figure*}[h]
  \includegraphics[width=0.94\linewidth]{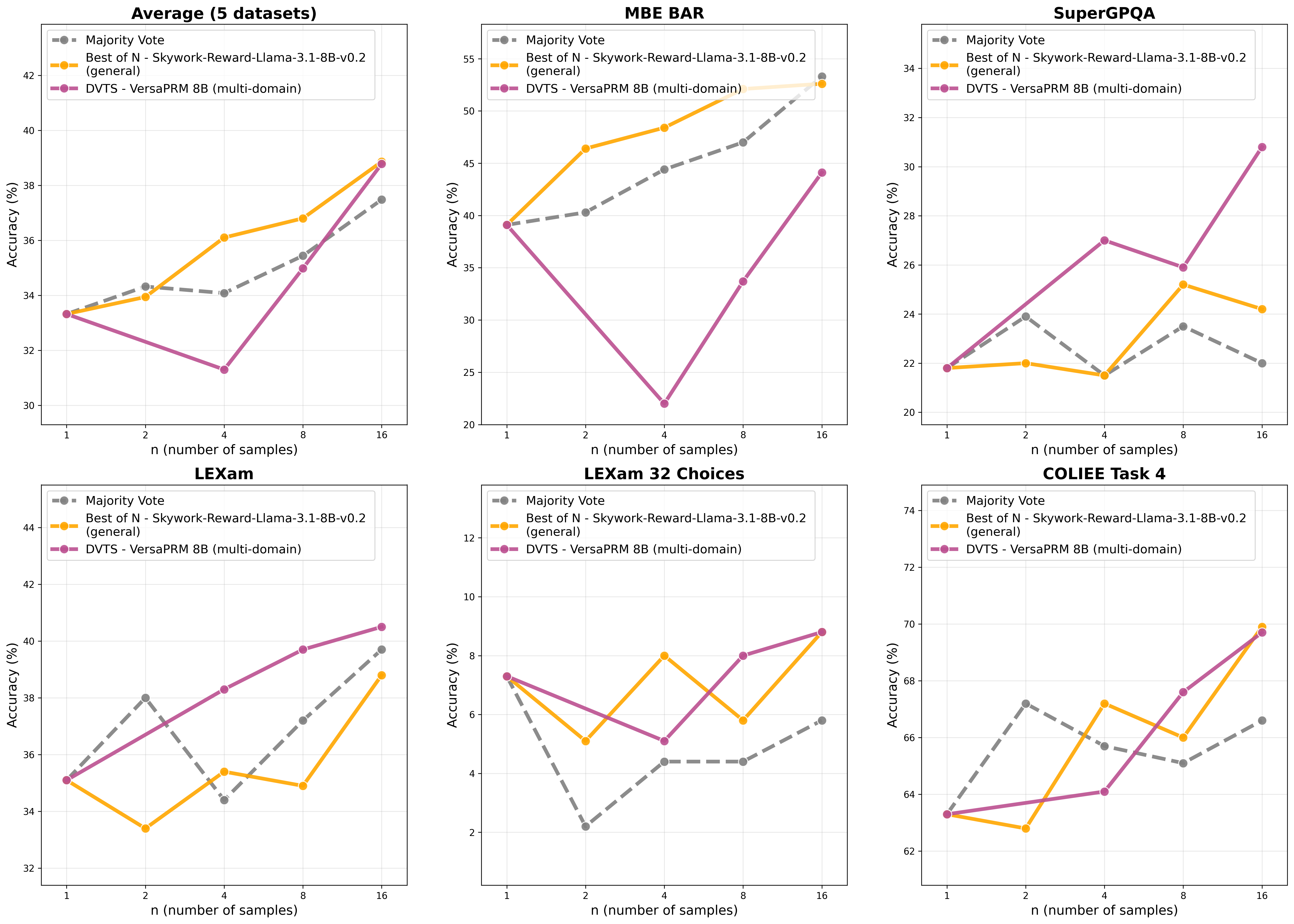}
  \caption {RQ1 average and individual benchmarks results using \texttt{Llama-3.2-3B-Instruct}}
  \label{fig:rq1-llama3}
\end{figure*}

\begin{figure*}[h]
  \includegraphics[width=0.94\linewidth]{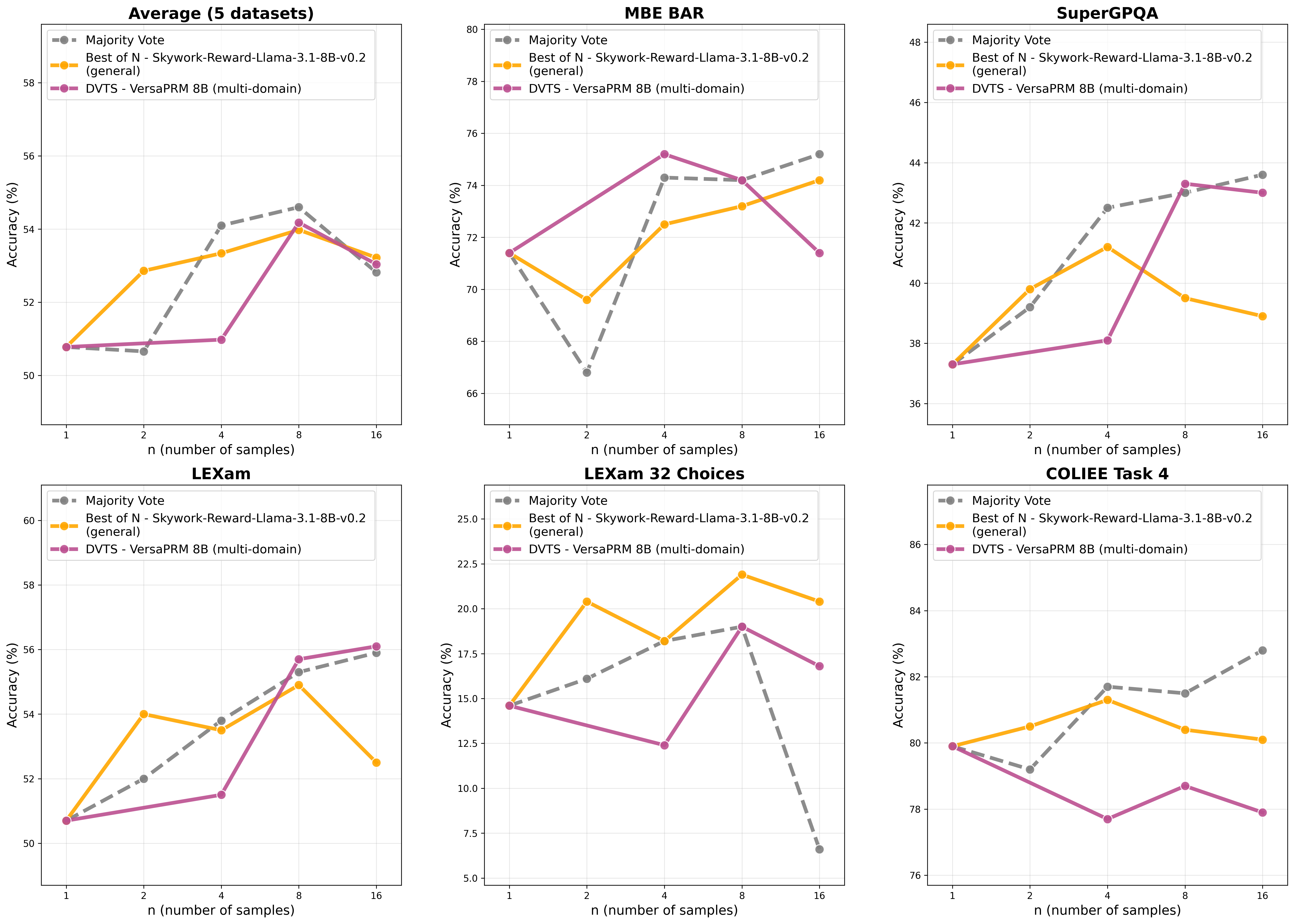}
  \caption {RQ1 average and individual benchmarks results using \texttt{Llama-3.1-70B-Instruct}}
  \label{fig:rq1-llama70b}
\end{figure*}

\begin{figure*}[h]
  \includegraphics[width=1\linewidth]{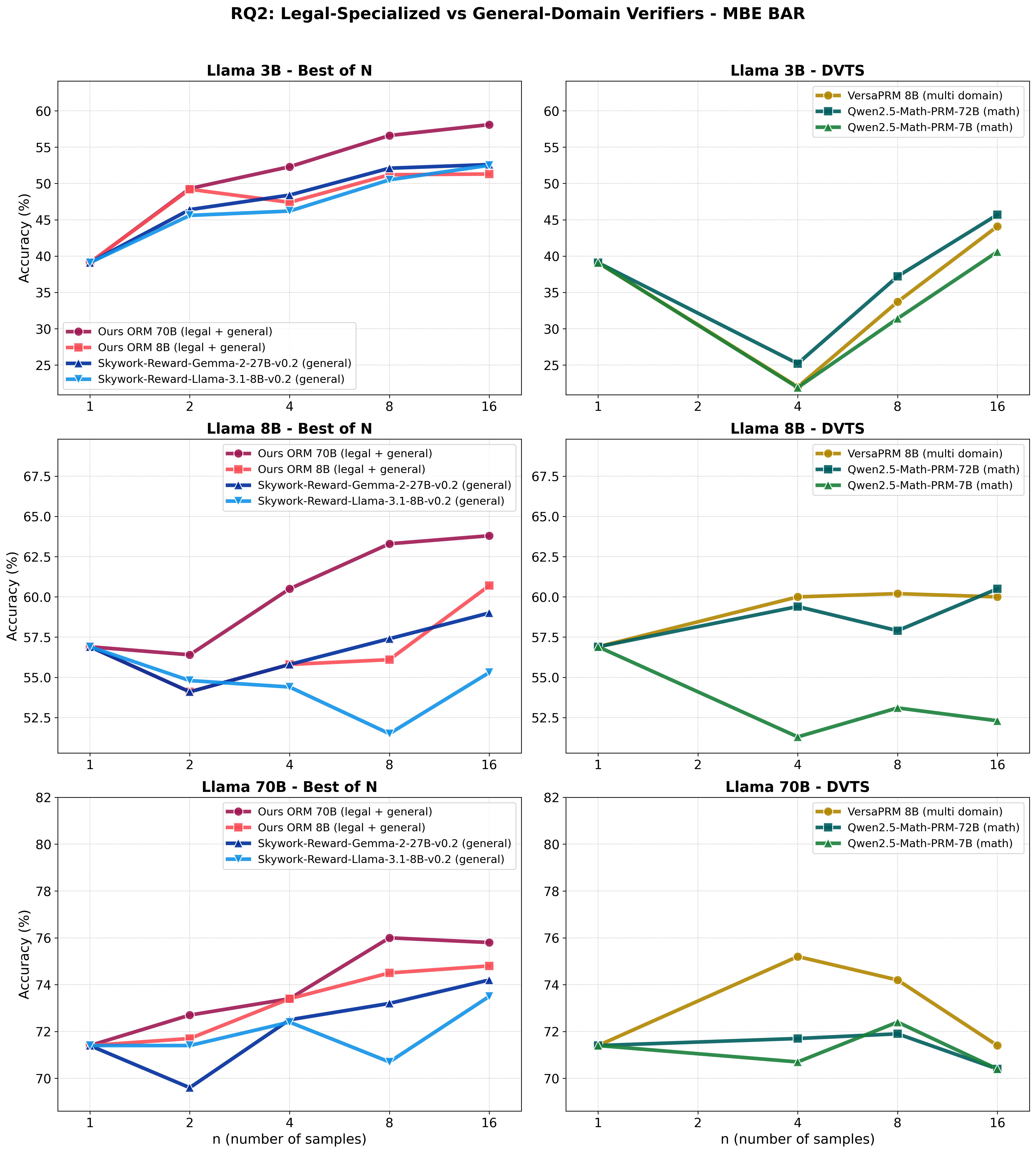}
  \caption{MBE bar exam RQ2 results with Best-of-N and DVTS}
  \label{fig:rq2_bar}
\end{figure*}
\begin{figure*}[h]
  \includegraphics[width=1\linewidth]{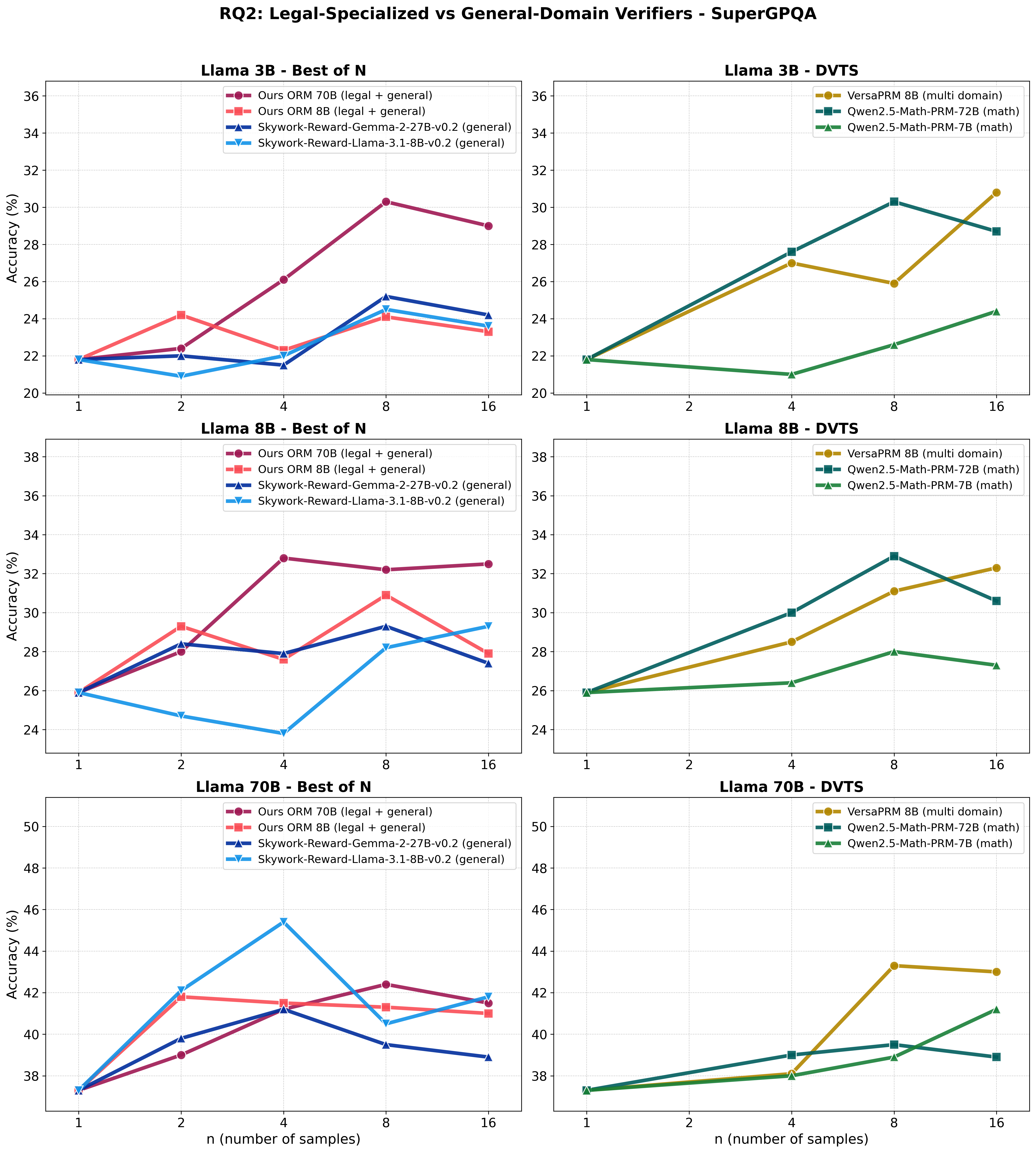}
  \caption{Coliee Task 4 RQ2 results with Best-of-N and DVTS}
  \label{fig:rq2_superg}
\end{figure*}
\begin{figure*}[h]
  \includegraphics[width=1\linewidth]{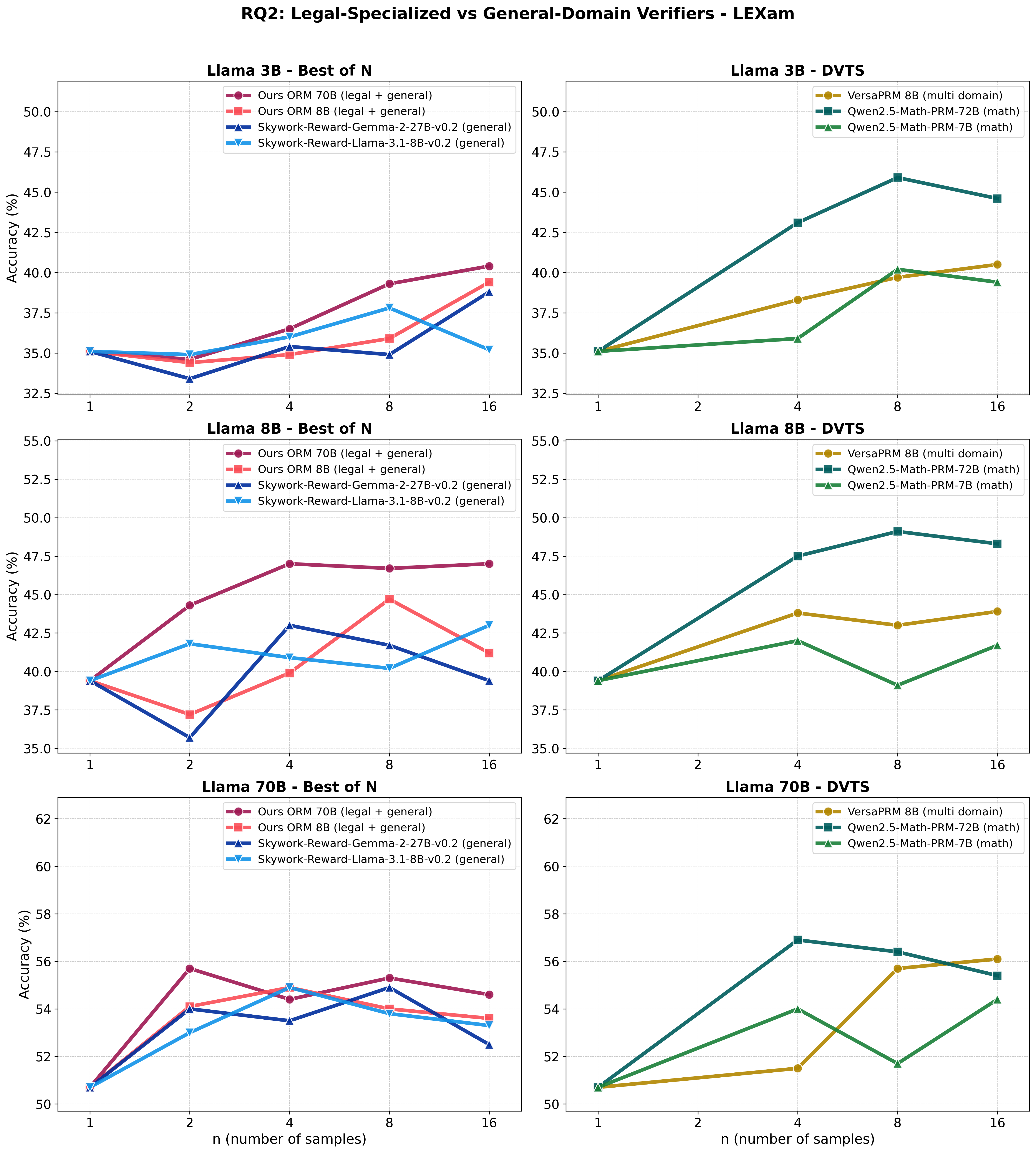}
  \caption{LEXam RQ2 results with Best-of-N and DVTS}
  \label{fig:rq2_lexam}
\end{figure*}
\begin{figure*}[h]
  \includegraphics[width=1\linewidth]{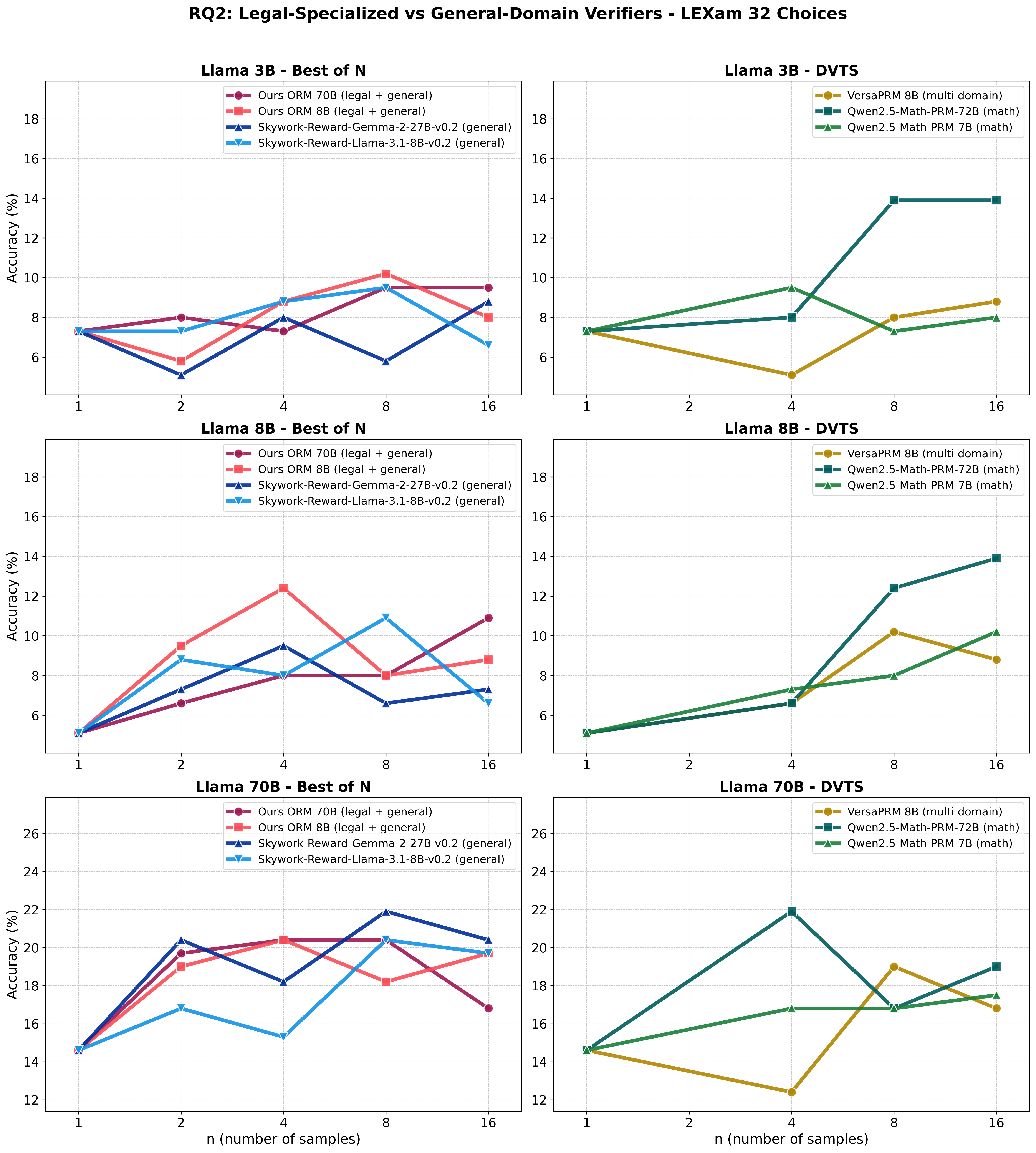}
  \caption{LEXam (32 options) RQ2 results with Best-of-N and DVTS}
  \label{fig:rq2_lexam32}
\end{figure*}
\begin{figure*}[h]
  \includegraphics[width=1\linewidth]{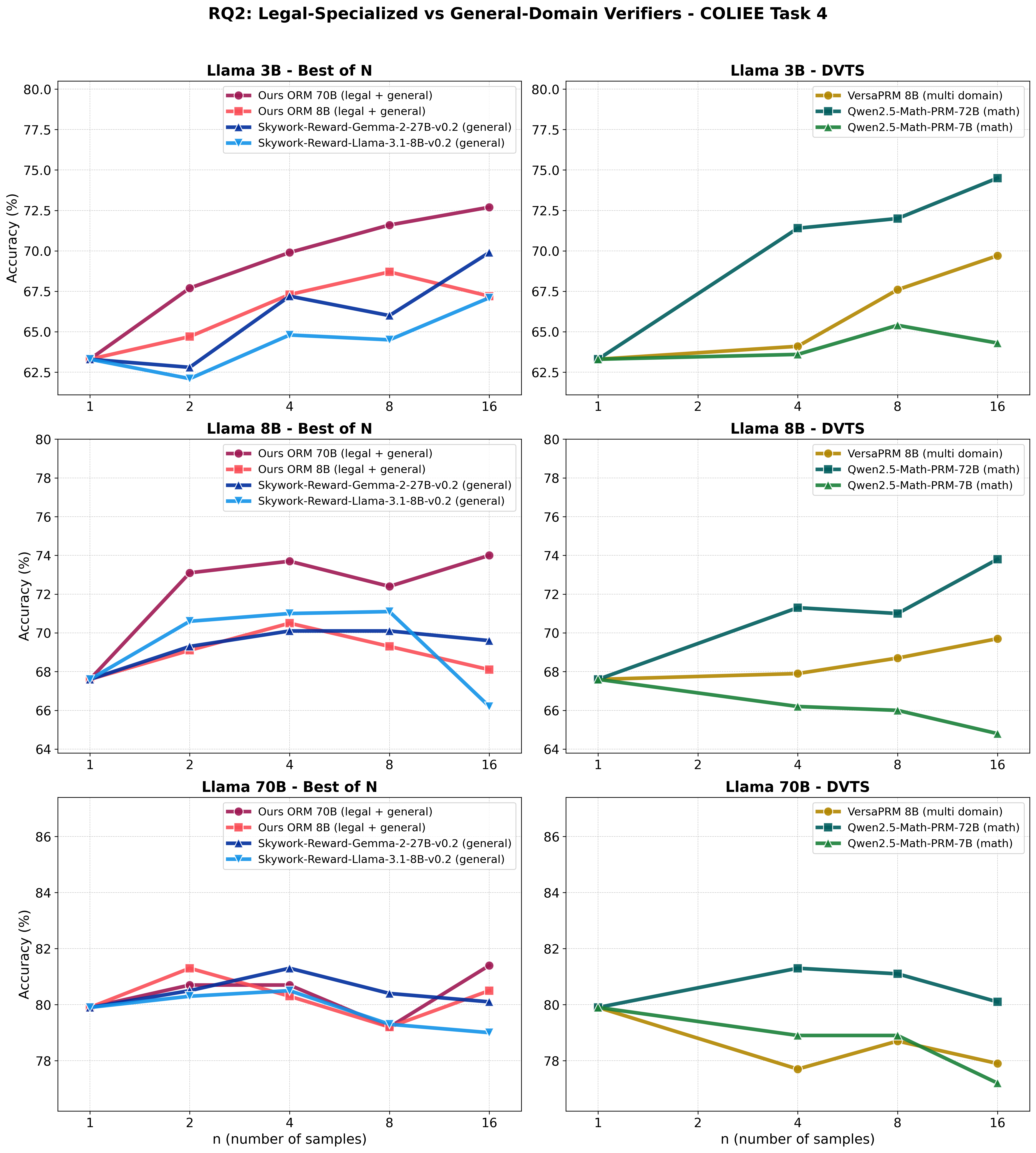}
  \caption{SuperGPQA RQ2 results with Best-of-N and DVTS}
  \label{fig:rq2_coliee}
\end{figure*}

\begin{figure*}[h]
\centering
  \includegraphics[width=0.9\linewidth]{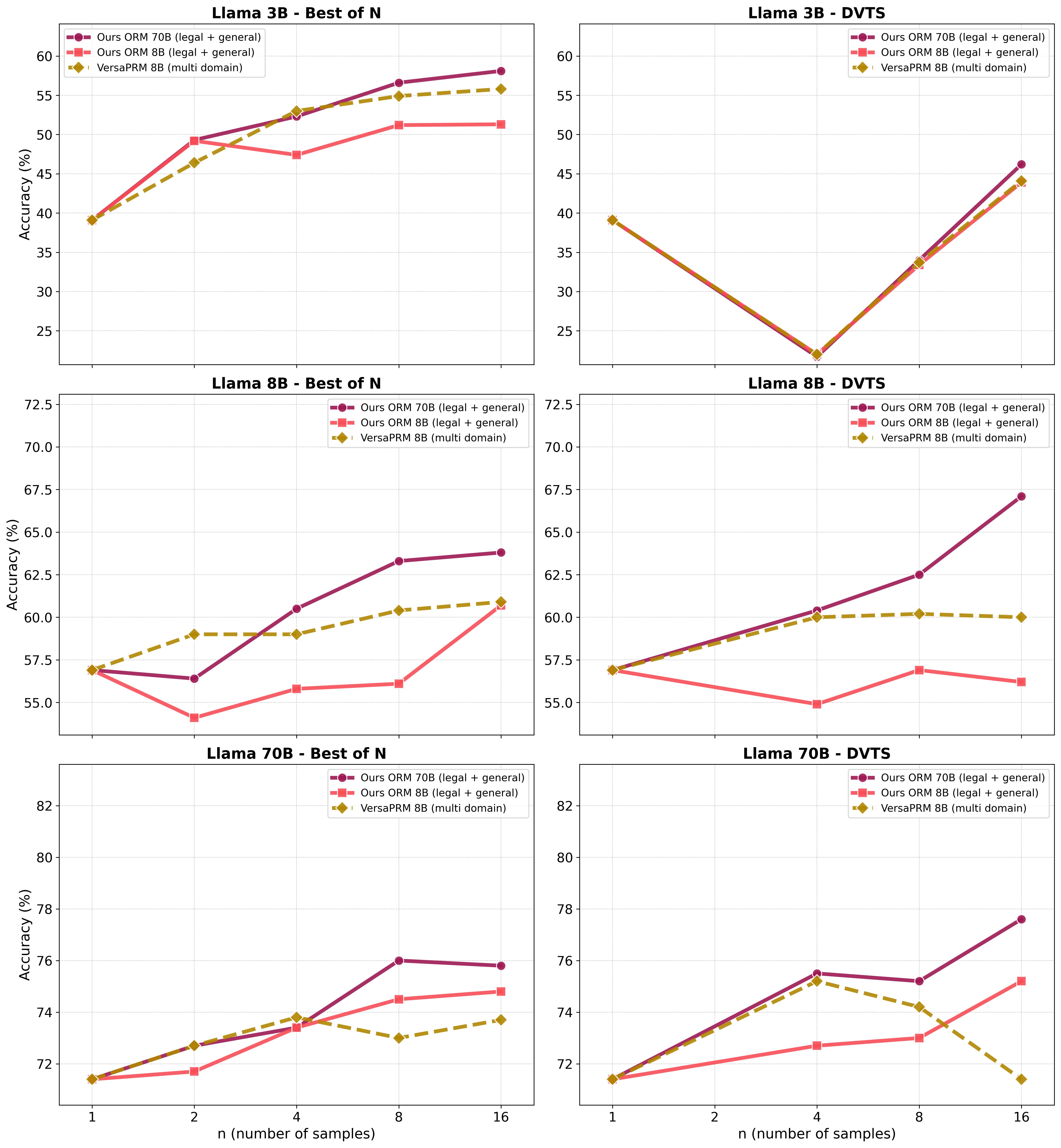}
  \caption{MBE bar exam RQ3 results with Best-of-N and DVTS}
  \label{fig:rq3_bar}
\end{figure*}
\begin{figure*}[h]
\centering
  \includegraphics[width=0.9\linewidth]{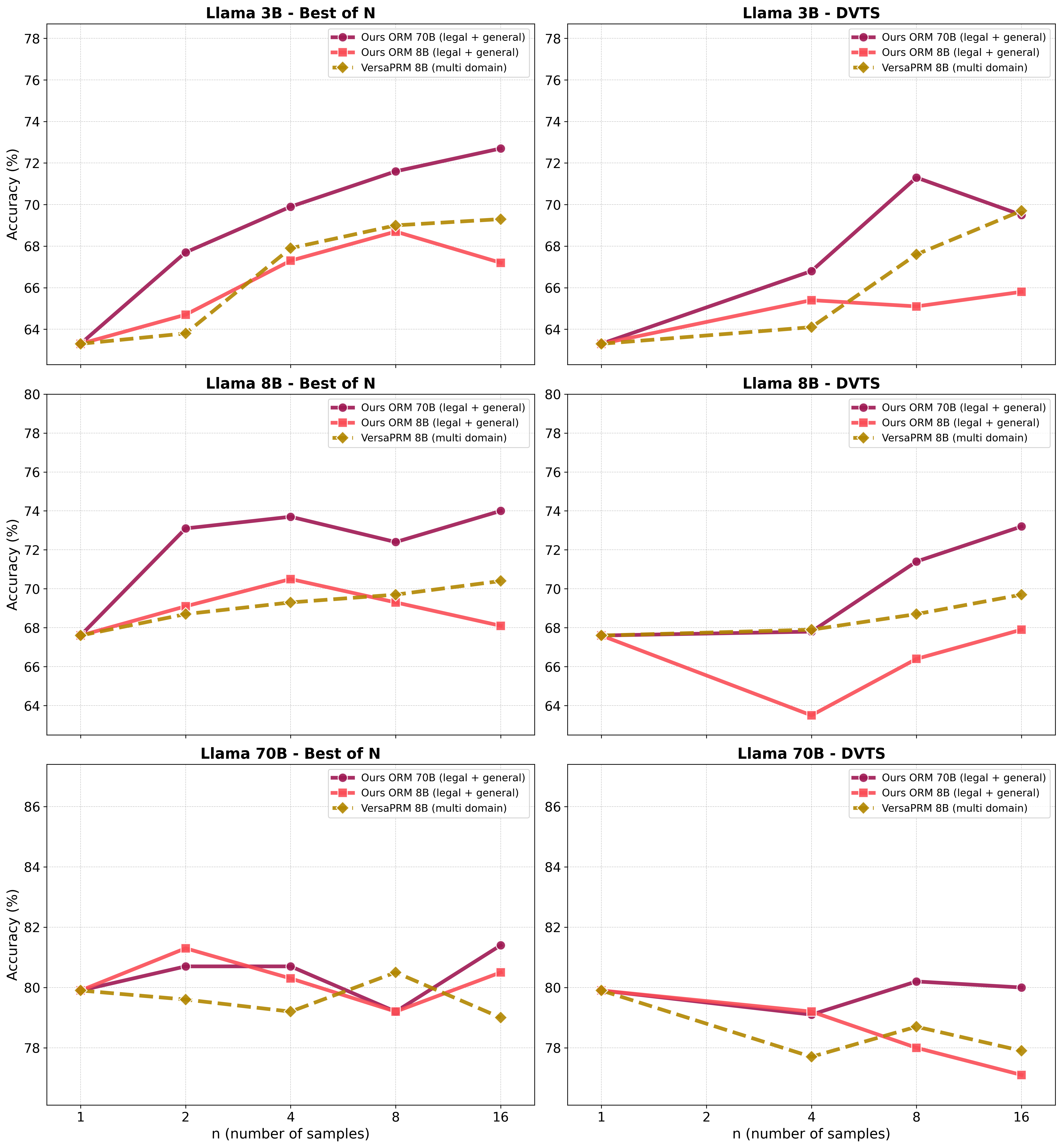}
  \caption{Coliee Task 4 RQ3 results with Best-of-N and DVTS}
  \label{fig:rq3_coliee4}
\end{figure*}
\begin{figure*}[h]
\centering
  \includegraphics[width=0.9\linewidth]{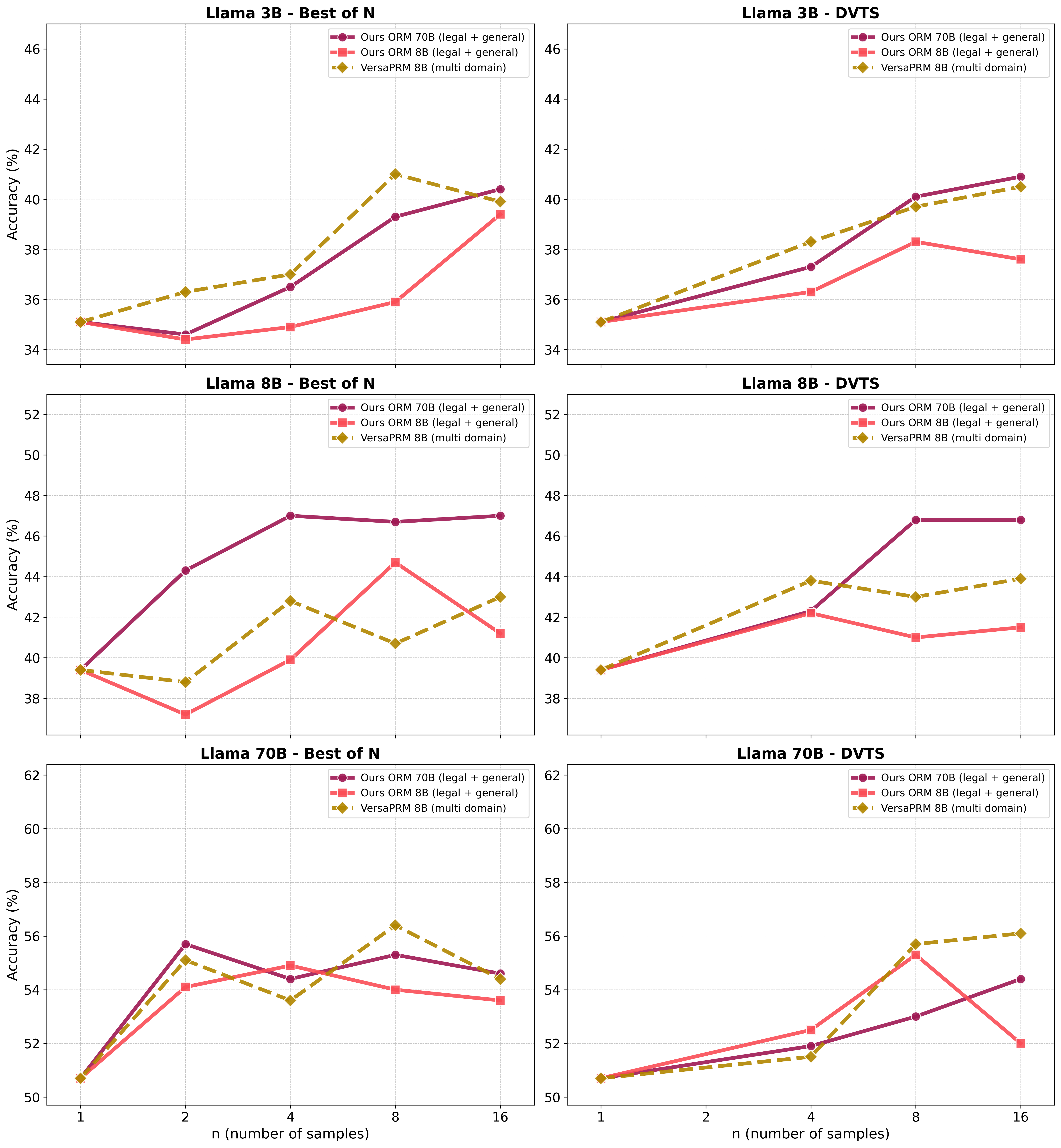}
  \caption{LEXam RQ3 results with Best-of-N and DVTS}
  \label{fig:rq3_lexam}
\end{figure*}
\begin{figure*}[h]
\centering
  \includegraphics[width=0.9\linewidth]{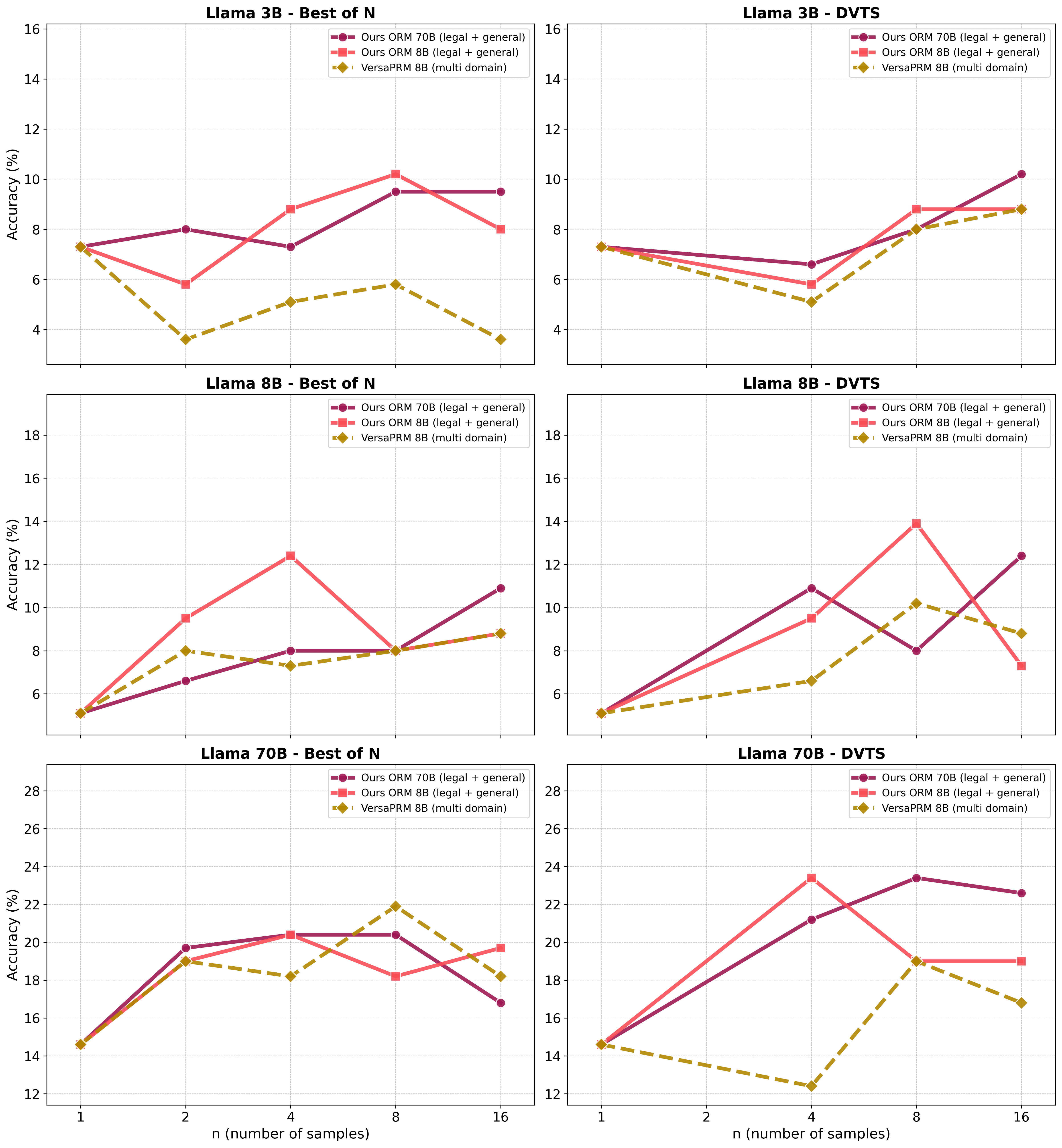}
  \caption{LEXam (32 options) RQ3 results with Best-of-N and DVTS}
  \label{fig:rq3_lexam32}
\end{figure*}

\begin{figure*}[h]
\centering
  \includegraphics[width=0.9\linewidth]{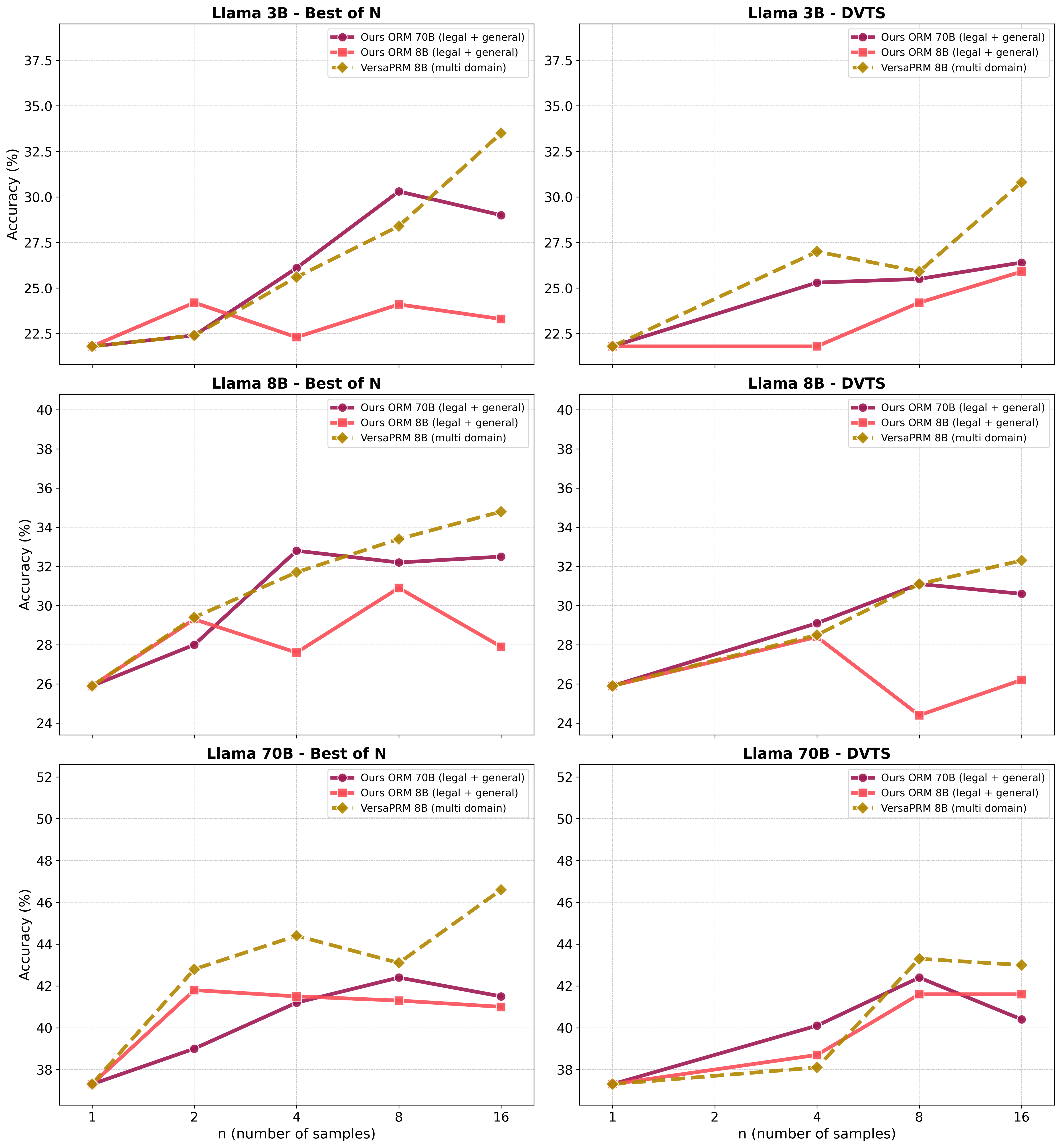}
  \caption{SuperGPQA RQ3 results with Best-of-N and DVTS}
  \label{fig:rq3_superg}
\end{figure*}


\end{document}